\documentclass[journal]{IEEEtran}

\usepackage{amssymb,amsmath,graphicx,times,cite,multirow,color,subfigure}
\usepackage[lined,ruled,boxed]{algorithm2e}

\begin{document}
\title{Transfer Learning for Brain-Computer Interfaces:\\ A Euclidean Space Data Alignment Approach}

\author{He~He and Dongrui~Wu
\thanks{He~He and Dongrui~Wu are with the Key Laboratory of Image Processing and Intelligent Control (Huazhong University of Science and Technology), Ministry of Education. They are also with the School of Artificial Intelligence and Automation, Huazhong University of Science and Technology, Wuhan, China. Email: hehe91@hust.edu.cn, drwu@hust.edu.cn.}
\thanks{Dongrui Wu is the corresponding author.}
\thanks{This work has been submitted to the IEEE for possible publication. Copyright may be transferred without notice, after which this version may no longer be accessible.}}

\maketitle

\begin{abstract}
\emph{Objective}: This paper targets a major challenge in developing practical EEG-based brain-computer interfaces (BCIs): how to cope with individual differences so that better learning performance can be obtained for a new subject, with minimum or even no subject-specific data?
\emph{Methods}: We propose a novel approach to align EEG trials from different subjects in the Euclidean space to make them more similar, and hence improve the learning performance for a new subject. Our approach has three desirable properties: 1) it aligns the EEG trials directly in the Euclidean space, and any signal processing, feature extraction and machine learning algorithms can then be applied to the aligned trials; 2) its computational cost is very low; and, 3) it is unsupervised and does not need any label information from the new subject.
\emph{Results}: Both offline and simulated online experiments on motor imagery classification and event-related potential classification verified that our proposed approach outperformed a state-of-the-art Riemannian space data alignment approach, and several approaches without data alignment.
\emph{Conclusion}: The proposed Euclidean space EEG data alignment approach can greatly facilitate transfer learning in BCIs.
\emph{Significance}: Our proposed approach is effective, efficient, and easy to implement. It could be an essential pre-processing step for EEG-based BCIs.
\end{abstract}


\IEEEpeerreviewmaketitle

\section{Introduction}

A brain-computer interface (BCI) \cite{Wolpaw2002,Lance2012} is a communication pathway for a user to interact with his/her surroundings by using brain signals, which contain information about the user's cognitive state or intentions. Electroencephalogram (EEG) is the most popular input in BCI systems. Motor imagery (MI) and event-related potentials (ERPs) are two common approaches of EEG-based BCIs, and also the focus of this paper.

For MI-based BCIs, the user needs to imagine the movements of his/her body parts (e.g., hands, feet, and tongue), which causes modulations of brain rhythms in the involved cortical areas. So, the imagination of different movements can be distinguished from the spatial localization of different sensorimotor rhythm modulations, and then used to control external devices. For ERP-based BCIs, the user is stimulated by a majority of common stimuli (non-target) and a small number of rare stimuli (target). The EEG response shows a special ERP pattern after the user perceives a target stimulus. So, a target stimulus can be detected by determining if there is an ERP pattern associated with it.

Early BCI systems were mainly used to help people with disabilities \cite{Pfurtscheller2008}. For example, MI-based BCIs have been used to help severely paralyzed patients to control powered exoskeletons or wheelchairs without the involvement of muscles, and ERP spellers enable patients who can not move nor speak to type. Recently, the application scope of BCIs has been extended to able-bodied people \cite{Nicolas-Alonso2012,Erp2012}, and EEG becomes the most popular input signal because it is easy and safe to acquire, and has high temporal resolution. However, EEG measures the very weak brain electrical signals from the scalp, which results in poor spatial resolution and low signal-to-noise ratio \cite{Blankertz2008}.

Consequently, sophisticated signal processing and machine learning algorithms are needed in EEG-based BCI systems to decode the EEG signal, especially for single-trial classification of EEG signals in real-world applications. Usually the EEG signals are first band-pass filtered and spatially filtered to increase the signal-to-noise ratio, and then discriminative features are extracted, which are next fed into machine learning algorithms such as Linear Discriminant Analysis (LDA) and Support Vector Machine (SVM) \cite{Bishop2006} for classification.

The covariance matrix of multi-channel EEG signals plays an important role in signal processing. For instance, common spatial pattern (CSP) filters \cite{Koles1990, Muller1999, Ramoser2000, drwuTLCSP2017}, computed directly from the covariance matrices, are the most popular spatial filters for MI. An intuitive explanation is that the interactions between different channels are encoded in the covariance matrices, which can be decomposed to find the spatial distribution of brain activities.

Recent years have also witnessed an increasing interest in using the EEG covariance matrices for both classification and regression \cite{Barachant2012, Congedo2013, Yger2017, drwuRG2017}. Since the covariance matrices are symmetric positive definite (SPD) and lie on a Riemannian manifold, a popular approach is to view each covariance matrix as a point in the Riemannian space, and use its geodesic to the Riemannian mean as a feature in classification. This approach is called the Minimum Distance to Riemannian Mean (MDRM) classifier \cite{Barachant2012, Congedo2013, Yger2017}.

MDRM can be directly applied to MI-based BCIs because the spatial information plays the most critical role in decoding MI signals. However, the discriminative information of ERP signals is represented temporally rather than spatially. So Barachant and Congedo \cite{Barachant2014} augmented the ERP trials to embed this temporal information. More specifically, the mean of the ERP trials is concatenated to each trial. The covariance matrix of the concatenated trial then contains both temporal and spatial information, which makes MDRM also applicable to ERP classification.

Transfer learning (TL) \cite{Pan2010}, which utilizes information in source domains to improve the learning performance in a target domain, has also been successfully used for BCIs \cite{Jayaram2016, drwuTNSRE2016, drwuTFS2017, drwuTHMS2017, drwuSMC2017}. Kang \emph{et al.} \cite{Kang2009} and Lotte and Guan \cite{Lotte2010} improved covariance matrix estimation for CSP filters by regularizing it towards the average of other subjects, or constructing a common feature space. Samek \emph{et al.} \cite{Samek2013} proposed an approach to transfer information about non-stationarities in the data to reduce the shift between subjects, and verified its performance in MI BCIs. Kindermans \emph{et al.} \cite{Kindermans2014a} integrated dynamic stopping, transfer learning and language model in a probabilistic zero-training framework and demonstrated competitive performance to a state-of-the-art supervised classifier in an ERP speller. Kobler and Scherer \cite{Kobler2016} pre-trained a Restricted Boltzmann Machine on a publicly available dataset and then adapted it to new observations in sensory motor rhythm based BCI.

Recently, Zanini \emph{et al.} \cite{Zanini2018} proposed a TL framework for the MDRM classifier, which is denoted as Riemannian alignment (RA)-MDRM in this paper, by utilizing the information of the resting state. In MI, the resting state is the time window that the subject is not performing any task, e.g., the transition window between two successive imageries. In ERP, particularly rapid serial visual presentation (RSVP), the stimuli are presented quickly one after another and the responses overlap, so it is difficult to find the resting state. \cite{Zanini2018} used the non-target stimuli as the resting state in ERP, which means some labeled data from the new subject must be known.

Experiments have shown that RA-MDRM outperformed MDRM in MI and ERP tasks \cite{Zanini2018}, when compared in a TL setting. But as mentioned above, it still needs a small amount of labeled subject-specific calibration trials for ERP classification. Moreover, for both MI and ERP, the classification is performed in the Riemannian space, whose geodesic computation is much more complicated, time-consuming, and unstable than the distance calculation in the Euclidean space. In this paper we propose a new EEG data alignment approach in the Euclidean space, which has the following desirable characteristics:
\begin{enumerate}
\item It transforms and aligns the EEG trials in the Euclidean space, and any signal processing, feature extraction and machine learning algorithms can then be applied to the aligned trials. On the contrary, RA aligns the covariance matrices (instead of the EEG trials themselves) in the Riemannian space, and hence a subsequent classifier must be able to operate on the covariance matrices directly, whereas there are very few such classifiers.
\item It can be computed several times faster than RA.
\item It only requires unlabeled EEG trials but does not need any label information from the new subject; so, it can be used in completely unsupervised learning.
\end{enumerate}
The effectiveness of our proposed approach is then demonstrated in two BCI classification scenarios:
\begin{enumerate}
  \item \textit{Offline unsupervised classification}, in which unlabeled EEG trials from a new subject are available, and we need to label them by making use of auxiliary labeled data from other subjects.

  \item \textit{Simulated online supervised classification}, in which a small number of labeled EEG epochs from a new subject are obtained sequentially on-the-fly, and a classifier is trained from them and auxiliary labeled data from other subjects to label future incoming epochs from the new subject.
\end{enumerate}

The remainder of this paper is organized as follows: Section~\ref{sect:RA-MDRM} introduces the RA-MDRM approach in the Riemannian space. Section~\ref{sect:method} proposes our Euclidean space data alignment approach. Section~\ref{sect:Datasets} introduces the three datasets used in our experiments, including two MI datasets and one ERP dataset. Sections~\ref{sect:Offline} and \ref{sect:Online} compare the performance of our approach with RA-MDRM in offline and simulated online learning, respectively. Finally, Section~\ref{sect:conclusions} draws conclusion and points out some future research directions.

\section{Related Work} \label{sect:RA-MDRM}

The covariance matrices of EEG trials are SPD, and lie in a Riemannian space instead of a Euclidean space \cite{Yger2017}. Since the covariance matrices directly encode the spatial information of the EEG trials, and by appropriately augmenting the EEG trials (such as in ERP classification) they can also encode the temporal information, we can perform EEG classification directly based on the covariance matrices.

This section introduces the MDRM classifier, which assigns a trial to the class whose Riemannian mean is the closest to its covariance matrix, and also a Riemannian space covariance matrix alignment approach (RA).

\subsection{Riemannian Distance}

The Riemannian distance between two SPD matrices $P_1$ and $P_2$ is called the \textit{geodesic}, which is the minimum length of a curve connecting them on the Riemannian manifold:
\begin{align}
 \delta(P_1,P_2)=\parallel \log(P_1^{-1}P_2)\parallel_F=\left[\sum_{r=1}^R\log^2\lambda_r\right]^{\frac{1}{2}}, \label{eq:geodesic}
\end{align}
where the subscript $F$ denotes the Frobenius norm, and $\lambda_r$ ($r=1,2,\cdots, R$) are the real eigenvalues of $P_1^{-1}P_2$.

The Riemannian distance between two SPD matrices $P_1$ and $P_2$ remains unchanged under linear invertible transformation:
\begin{align}
 \delta(C^TP_1C,C^TP_2C)= \delta(P_1,P_2),\label{eq:congruence invariance}
\end{align}
where $C$ is an invertible matrix. This property of the Riemannian distance is called \emph{congruence invariance}.

\subsection{Riemannian Mean}

The mean of a set of SPD matrices can be computed in the Euclidean space as their arithmetic mean, and also in the Riemannian space as the Riemannian mean (geometric mean), defined as the matrix minimizing the sum of the squared Riemannian distances:
\begin{align}
  \varrho(P_1,\cdots,P_N)=\arg\min_{P}\sum_{n=1}^N\delta^2(P,P_n). \label{eq:RGmean}
\end{align}
There is no closed-form solution to (\ref{eq:RGmean}), and it is usually computed by an iterative gradient descent algorithm \cite{Fletcher2004}.

\subsection{MDRM} \label{sect:MDRM0}

The MDRM classifier \cite{Barachant2012, Congedo2013, Yger2017} first computes the Riemannian mean of each class from the covariance matrices of the labeled training trials, then assigns each test trial to the class whose Riemannian mean is the closest to its covariance matrix, i.e.,
\begin{align}\label{eq:mdrm}
  g(\Sigma)=\arg\min_{c}\delta(\Sigma,\bar\Sigma^c),
\end{align}
where $\Sigma$ is the covariance matrix of the test trial, $\bar\Sigma^c$ is the Riemannian mean of Class~$c$, and $g(\Sigma)$ is the predicted class label of $\Sigma$.

\subsection{RA-MDRM} \label{sect:MDRM}

Zanini \emph{et al.} \cite{Zanini2018} proposed a novel TL approach in the Riemannian space, referred to in this paper as RA-MDRM, to improve the performance of the MDRM classifier by utilizing auxiliary data from other sessions and/or subjects when there are only a few labeled trials from a new subject. Since the covariance matrices of the trials are the input to MDRM, RA-MDRM aims to align the covariance matrices from different sessions/subjects to give them a common reference. \cite{Zanini2018} assumes that ``\emph{different source configurations and electrode positions induce shifts of covariance matrices with respect to a reference
(resting) state, but that when the brain is engaged in a specific task, covariance matrices move over the SPD manifold in the same direction.}" Then RA-MDRM centers ``\emph{the covariance matrices of every session/subject with respect to a reference covariance matrix so that what we observe is only the displacement with respect to the reference state due to the task.}"

More specifically, RA-MDRM first computes the covariance matrices of some \emph{resting} trials, $\{R_i\}_{i=1}^k$, in which the subject is not performing any task, and then computes the Riemannian mean $\bar{R}$ of these matrices. $\bar{R}$ is then used as the reference matrix in RA-MDRM to reduce the inter-session/subject variability by the following transformation:
\begin{align}
  \tilde{\Sigma}_i=\bar{R}^{-1/2}\Sigma_i\bar{R}^{-1/2}, \label{eq:15}
\end{align}
where $\Sigma_i$ is the covariance matrix of the $i$th trial, and $\tilde{\Sigma}_i$ is the corresponding aligned covariance matrix.

Equation (\ref{eq:15}) makes the reference state of different sessions/subjects centered at the identity matrix. This transformation would not change the distance between the covariance matrices belonging to the same session/subject because of the congruence invariance property in (\ref{eq:congruence invariance}), but makes the covariance matrices of different sessions/subjects move over the Riemannian manifold in different directions with respect to the corresponding reference matrices, and hence reduces the cross-session/subject differences. As a result, covariance matrices from different sessions/subjects can be aligned and become comparable if $\bar{R}$ can be appropriately estimated.

In MI, the resting state is the time window that the subject is not performing any task, e.g., the transition window between two imageries. In ERP, particularly the RSVP, the stimuli are presented quickly one after another and the responses overlap, so it is difficult to find the resting state. \cite{Zanini2018} used the non-target stimuli as the resting state in ERP, which requires that some labeled trials from the new subject must be known. That is, in ERP,
\begin{align}
  \bar{R}=\arg\min_{R}\sum_{i\in I}\delta^2(R, \Sigma_i), \label{eq:ERPref}
\end{align}
where $I$ is the index set of the non-target trials.

RA-MDRM can be applied to both MI and ERP data; however, there is an important difference in building covariance matrices in these two paradigms.

Specifically, the covariance matrix of an MI trial $X_i$ is simply computed as:
\begin{align}
 \Sigma_i=X_iX_i^T. \label{eq:Sigma}
\end{align}
$\Sigma_i$ encodes the most discriminative information of an MI trial, i.e., the spatial distribution of the brain activity.

However, the main discriminative information of ERP trials is carried temporally rather than spatially. The normal covariance matrix such as (\ref{eq:Sigma}) ignores this temporal information. So Barachant and Congedo \cite{Barachant2014} proposed a novel approach to augment the ERP trials so that their covariance matrices can also encode the temporal information. They first compute the mean of the ERP trials:
\begin{align}
 \bar{X}=\frac{1}{\mid I\mid}\sum_{i\in I} X_i,
\end{align}
where $I$ is the index set of the ERP trials. They then build an augmented trial $X_i^*$ by concatenating $\bar{X}$ and $X_i$:
\begin{align}
 X_i^*=\begin{bmatrix}\bar{X}\\X_i \end{bmatrix}\label{eq:Xi}
\end{align}
The covariance matrix of $X_i^*$ is then used in RA-MDRM.

\subsection{Limitations of RA}

Although RA-MDRM has demonstrated promising performance in several BCI applications \cite{Zanini2018}, it still has some limitations:
\begin{enumerate}
  \item RA-MDRM aligns the covariance matrices in the Riemannian space, instead of the EEG trials themselves. A subsequent classifier must be able to operate on the covariance matrices directly, whereas there are very few such classifiers.
  \item RA-MDRM uses the Riemannian mean of the covariance matrices, which is time-consuming to compute, especially when the number of EEG channels is large.
  \item RA-MDRM for ERP classification needs some labeled trials from the new subject, more specifically, RA needs some non-target trials to compute the reference matrix in (\ref{eq:ERPref}), and MDRM needs some target trials to construct $X_i^*$ in (\ref{eq:Xi}), so it is a supervised learning approach and cannot be used when there is no label information from the new subject at all.
\end{enumerate}

\section{EEG Data Alignment in the Euclidean Space (EA)} \label{sect:method}

This section introduces our proposed Euclidean-space alignment (EA) approach.

\subsection{The EA}

To cope with the limitations of RA, we propose EA that does not need any labeled data from the new subject, and can be computed much more efficiently. The rationale is to make the data distributions from different subjects more similar, and hence a classifier trained on the auxiliary data would have a better chance to perform well on the new subject. This idea has been widely used in TL \cite{drwuTNSRE2016,Pan2010,Sun2016}.

Similar to RA, our approach is also based on a reference matrix $\bar{R}$, but estimated in a different way. Assume a subject has $n$ trials. Then,
\begin{align}
  \bar{R}=\frac{1}{n}\sum_{i=1}^nX_iX_i^T, \label{eq:ref2}
\end{align}
i.e., $\bar{R}$ is the arithmetic mean of all covariance matrices from a subject. We then perform the alignment by
\begin{align}
  \tilde{X}_i=\bar{R}^{-1/2}X_i. \label{eq:EA}
\end{align}
After the alignment, the mean covariance matrix of all $n$ aligned trials is:
\begin{align}
\frac{1}{n}\sum_{i=1}^n\tilde{X}_i\tilde{X}_i^T  &=\frac{1}{n}\sum_{i=1}^n\bar{R}^{-1/2}X_iX_i^T\bar{R}^{-1/2}\nonumber \\
&=\bar{R}^{-1/2}\left(\frac{1}{n}\sum_{i=1}^nX_iX_i^T\right)\bar{R}^{-1/2}\nonumber\\
&=\bar{R}^{-1/2}\bar{R}\bar{R}^{-1/2}=I, \label{eq:meanC}
\end{align}
i.e., the mean covariance matrices of all subjects are equal to the identity matrix after alignment, and hence the distributions of the covariance matrices from different subjects are more similar. This is very desirable in TL.

The idea of EA can also be explained using the concept of maximum mean discrepancy (MMD) \cite{Gretton2007,drwuTNSRE2016}, widely used in TL. MMD represents the distances between different distributions as the distances between their mean embeddings of features. Smaller distances indicate that the distributions are more similar, and hence more suitable for TL. If we view the covariance matrices as the feature embeddings of EEG trials, then, after EA, the distances between EEG trials from different subjects become zero (because the mean covariance matrices of all subjects are identical), which should generally benefit TL.

\subsection{Comparison with RA}

Both EA and RA ensure the Riemannian distances among the covariance matrices are kept unchanged after the alignment. However, there are three major differences between them:
\begin{enumerate}
\item RA computes the reference matrix $\bar{R}$ as the \emph{Riemannian (geometric) mean} of the \emph{resting state covariance matrices}, whereas EA computes the reference matrix $\bar{R}$ as the \emph{Euclidean (arithmetic) mean} of \emph{all covariance matrices}.
\item RA aligns the \emph{covariance matrices} in the \emph{Riemannian space}, whereas EA aligns the \emph{time domain EEG trials} in the \emph{Euclidean space}.
\item After RA, the \emph{Riemannian mean of the resting state covariance matrices} becomes an identity matrix (but the Riemannian mean of \emph{all} covariance matrices is not). After EA, the \emph{Euclidean mean of all covariance matrix} becomes an identity matrix.
\end{enumerate}

Compared with RA, EA has the following desirable properties:
\begin{enumerate}
\item EA transforms and aligns the EEG trials in the Euclidean space. Any subsequent signal processing, feature extraction and machine learning algorithms can then be applied to the aligned trials. So, it has much broader applications than RA, which aligns the covariance matrices (instead of the EEG trials) in the Riemannian space.
\item EA can be computed much faster than RA, because EA uses the arithmetic mean as the reference matrix, whereas RA uses the Riemannian mean as the reference matrix.
\item EA does not need any label information from the new subject, whereas RA needs some label information for ERP classification.
\end{enumerate}

\subsection{Relationship to CORAL} \label{sect:CORAL}

A ``\emph{frustratingly easy domain adaptation}" approach, CORrelation ALignment (CORAL) \cite{Sun2016}, was proposed in 2016 to minimize domain shift by aligning the second-order statistics of different distributions, without requiring any target labels. Its idea is very similar to EA.

CORAL considers 1D features (vectors), instead of 2D features (matrices) such as EEG trials in this paper. Let $C_S\in\mathbb{R}^{d_S\times d_S}$ and $C_T\in\mathbb{R}^{d_T\times d_T}$ be the feature covariance matrices in the source and target domains, respectively, where $d_S$ and $d_T$ are the number of features in the source and target domains, respectively. Then, CORAL finds a linear transformation $A\in\mathbb{R}^{d_S\times d_T}$ to the source domain features, so that the Frobenius norm of the difference between their covariance matrices is minimized, i.e.,
\begin{align}
\min_A||A^TC_SA-C_T||_F^2
\end{align}
The linear transformation $A$ has a simple closed-form solution \cite{Sun2016}.

EA and CORAL are similar; however, there are also some important differences:
\begin{enumerate}
\item CORAL considers \emph{1D} features, and each domain has only one covariance matrix, which measures the covariances between different pairs of individual features. EA considers \emph{2D} features (EEG trials), and each domain has many covariance matrices (each corresponding to one EEG trial), each of which measures the covariances between different pairs of EEG channels in an EEG trial.
\item CORAL minimizes the distance between the covariance matrices in different domains, whereas EA minimizes the distance between the \emph{mean} covariance matrices in different domains.
\item CORAL finds a linear transformation to the source domain features only, so that the transformed source domain covariance matrix approaches the original target domain covariance matrix. EA finds a separate linear transformation to each domain, so that the mean of the transformed source domain covariance matrices equals the mean of the transformed target domain covariance matrices.
\end{enumerate}

\section{Datasets} \label{sect:Datasets}

This section introduces two MI datasets and one ERP dataset used in our experiments.

\subsection{MI Datasets}

Two MI datasets from BCI Competition IV\footnote{http://www.bbci.de/competition/iv/.} were used. Their experimental paradigms were similar: In each session a subject sat in a comfortable chair in front of a computer. At the beginning of a trial, a fixation cross appeared on the black screen to prompt the subject to be prepared. A moment later, an arrow pointing to a certain direction was presented as a visual cue for a few seconds. In this period the subject was asked to perform a specific MI task without feedback according to the direction of the arrow. Then the visual cue disappeared from the screen and a short break followed until the next trial began.

The first dataset\footnote{http://www.bbci.de/competition/iv/desc\_1.html.} (Dataset~1 \cite{Blankertz2007}) was recorded from seven healthy subjects. For each subject two classes of MI were selected from three classes: left hand, right hand, and foot. Continuous 59-channel EEG signals were acquired for three phases: calibration, evaluation, and special feature. Here we only used the calibration data which provided complete marker information. Each subject had 100 trials from each class in the calibration phase.

The second MI dataset\footnote{http://www.bbci.de/competition/iv/desc\_2a.pdf.} (Dataset~2a) consisted of EEG data from nine heathy subjects. Each subject was instructed to perform four different MI tasks, namely the imagination of the movement of the left hand, right hand, both feet, and tongue. 22-channel EEG signals and 3-channel EOG signals were recorded at 250Hz. A training phase and an evaluation phase were recorded on different days for each subject. Here we only used the EEG data from the training phase, which included complete marker information. Additionally, two MI classes (left hand and right hand) were selected and each class had 72 trials.

A causal band-pass filter (50-order linear phase Hamming window FIR filter designed by Matlab function \emph{fir1}, with 6dB cut-off frequencies at [8, 30] Hz) was applied to remove muscle artifacts, line-noise contamination and DC drift. Next, we extracted EEG signals between $[0.5, 3.5]$ seconds after the cue appearance as our trials for both datasets. EEG signals between $[4.25, 5.25]$ seconds after the cue appearance were extracted as resting states.

\subsection{ERP Dataset}

We used an RSVP dataset from PhysioNet\footnote{https://www.physionet.org/physiobank/database/ltrsvp/.} \cite{Goldberger2000} for ERP classification. It contained EEG data from 11 healthy subjects upon rapid presentation of images at 5, 6, and 10 Hz \cite{Matran-Fernandez2017}. Each subject was seated in front of a computer showing a series of images rapidly. The images were aerial pictures of London falling into two categories, namely target images and non-target images. Target images contained a randomly rotated and positioned airplane that had been photo realistically superimposed, and non-target images did not contain airplanes. The task was to recognize if the images were target or non-target from EEG signals, which were recorded from 8 channels at 2048 Hz.

For each presentation rate and subject there were two sessions represented by ``a" and ``b", which indicated whether the first image was `` target" or ``non-target", respectively. Here we used the 5 Hz version (five images per second) in Session~a. The number of samples for different subjects varying between 368 and 565, and the target to non-target ratio was around 1:9.

The continuous EEG data had been bandpass filtered between $[0.15,28]$ Hz. We downsampled the EEG signal from 2048Hz to 64Hz, and epoched each trial to the $[0,0.7]$ second interval time-locked to the stimulus onset.

\subsection{Data Visualization} \label{sect:visualization}

\begin{figure}[htpb] \centering
\includegraphics[width=\linewidth,clip]{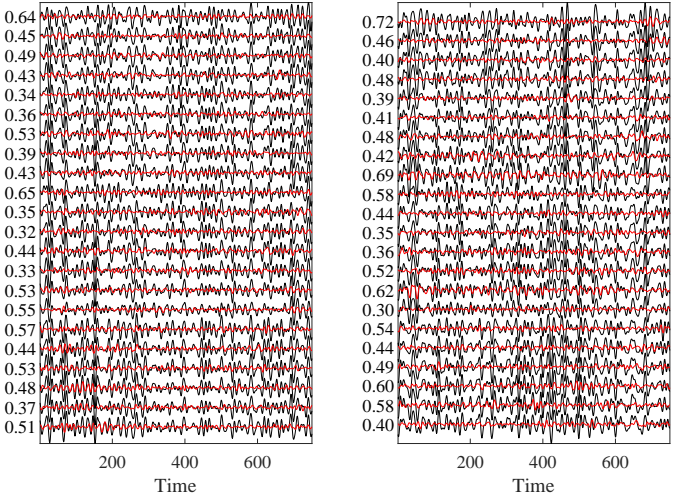}
\caption{EEG trials before (black curves) and after (red curves) EA. Each row is a different channel.} \label{fig:EA}
\end{figure}

It's interesting to visualize how the EEG trials are modified by EA. Fig.~\ref{fig:EA} shows two examples (one for left hand imagery, and the other for right) from Subject~1 in Dataset~2a. The black and red curves are EEG signals before and after EA, respectively, and the vertical axis numbers show their correlations. The magnitudes of the EEG signals are smaller and more uniform after EA, and the EEG signals before and after EA generally have low correlation.

To visualize how EA reduces individual differences, we used $t$-Stochastic Neighbor Embedding ($t$-SNE) \cite{Maaten2008}, a nonlinear dimensionality reduction technique that embeds high-dimensional data in a two- or three- dimensional space, to show and compare the EEG trials before and after EA.

\begin{figure}[htpb] \centering
\subfigure[]{\label{fig:MI1tSNE}     \includegraphics[width=.94\linewidth,clip]{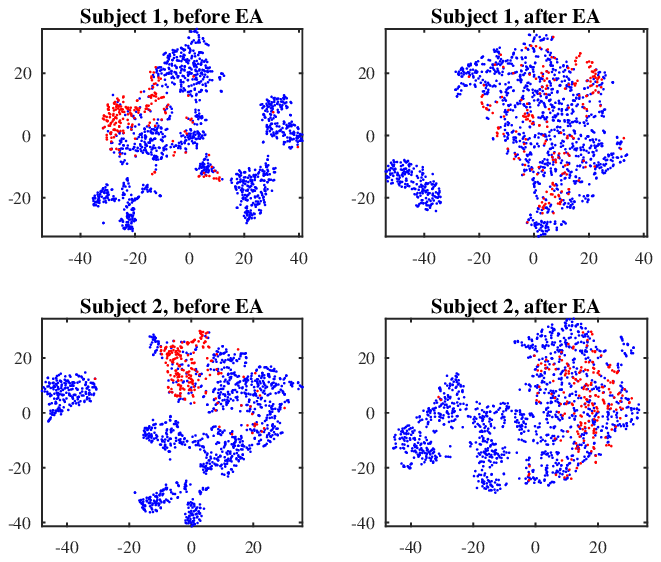}}
\subfigure[]{\label{fig:MI2tSNE}     \includegraphics[width=.94\linewidth,clip]{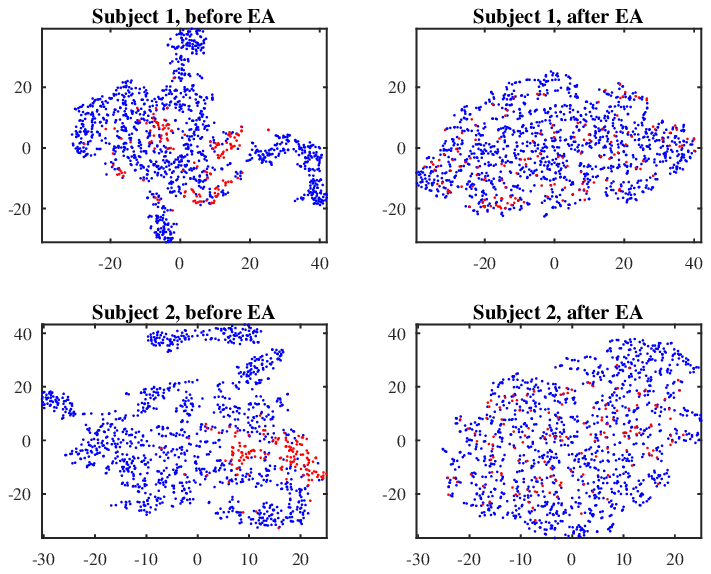}}
\subfigure[]{\label{fig:RSVPtSNE}    \includegraphics[width=.94\linewidth,clip]{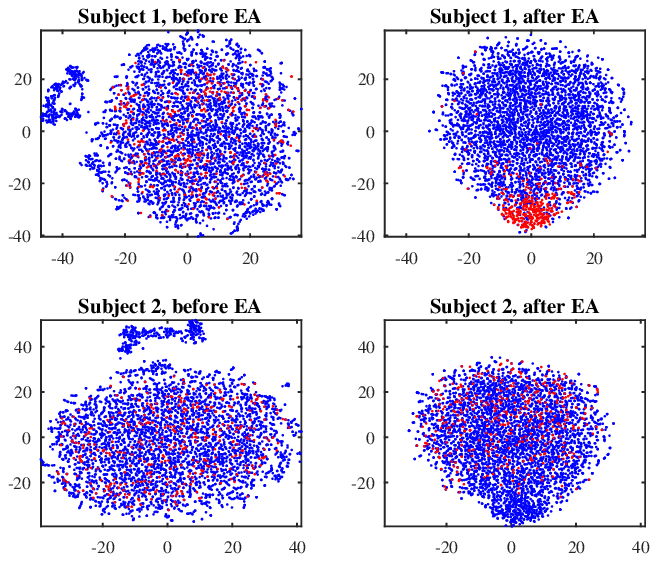}}
\caption{$t$-SNE visualization of the first two subjects before and after EA. (a) MI Dataset~1; (b) MI Dataset~2a; (c) ERP. Red dots: trials from the test subject; blue dots: trials from the training subjects.} \label{fig:tSNE}
\end{figure}

Each time we picked trials from one subject as the test set, and combined trials from all remaining subjects as the training set. Fig.~\ref{fig:MI1tSNE} shows the $t$-SNE visualization of the first two subjects in MI Dataset~1, each row corresponding to a different test subject. The red dots are trials from the test subject, and the blue dots from the training subjects. In each row, the left plot shows the trials before EA, and the right after EA. Corresponding visualization results for the first two subjects in MI Dataset~2a and ERP are shown in Figs.~\ref{fig:MI2tSNE} and \ref{fig:RSVPtSNE}, respectively.

The training trials (blue dots) may be scattered far away from the test trials (red dots) before EA, especially in Fig.~\ref{fig:MI1tSNE}. So, applying a classifier designed on the training trials directly to the test trials may not achieve good performance. However, after EA, the training and test trials overlap with each other, i.e., the discrepancies between them are reduced.

\section{Performance Evaluation: Offline Unsupervised Classification} \label{sect:Offline}

This section presents the performance comparison of EA with other approaches on both MI and ERP datasets in offline unsupervised classification.

\subsection{Offline Unsupervised Classification}

In each dataset, there were multiple subjects, and each subject was first aligned independently, either in the Riemannian space using (\ref{eq:15}), or in the Euclidean space using (\ref{eq:EA}). Since we had access to all EEG recordings in offline classification, all trials or resting epochs between all trials were used to estimate the reference matrices. We then used leave-one-subject-out cross-validation to evaluate the classification performance: each time we picked one subject as the new subject (test set), combined EEG trails from all remaining subjects as the training set to build the classifier, and then tested the classifier on the new subject.

\subsection{Offline Classification Results on the MI Datasets} \label{sect:OfflineMI}

We first tested EA on the two MI datasets, and compared its performance with RA-MDRM. In the Euclidean space, after EA, we used CSP \cite{Koles1990, Muller1999, Ramoser2000, drwuTLCSP2017} for spatial filtering and LDA for classification. More specifically, the following four approaches were compared:
\begin{enumerate}
  \item MDRM: The basic MDRM classifier, as introduced in Section~\ref{sect:MDRM0}. It does not include any data alignment.
  \item RA-MDRM: It is the approach introduced in Section~\ref{sect:MDRM}, which first aligns the covariance matrices in the Riemannian space, and then performs MDRM.
  \item CSP-LDA: It is a standard Euclidean space classification approach for MI, which spatially filters the EEG trials by CSP and then classifies them by LDA. It does not include any data alignment.
  \item EA-CSP-LDA: It first aligns the EEG trials in the Euclidean space by EA (Section~\ref{sect:method}), and then performs CSP filtering and LDA classification.
\end{enumerate}

\begin{figure}[htpb]\centering
\subfigure[]{\includegraphics[width=.8\linewidth,clip]{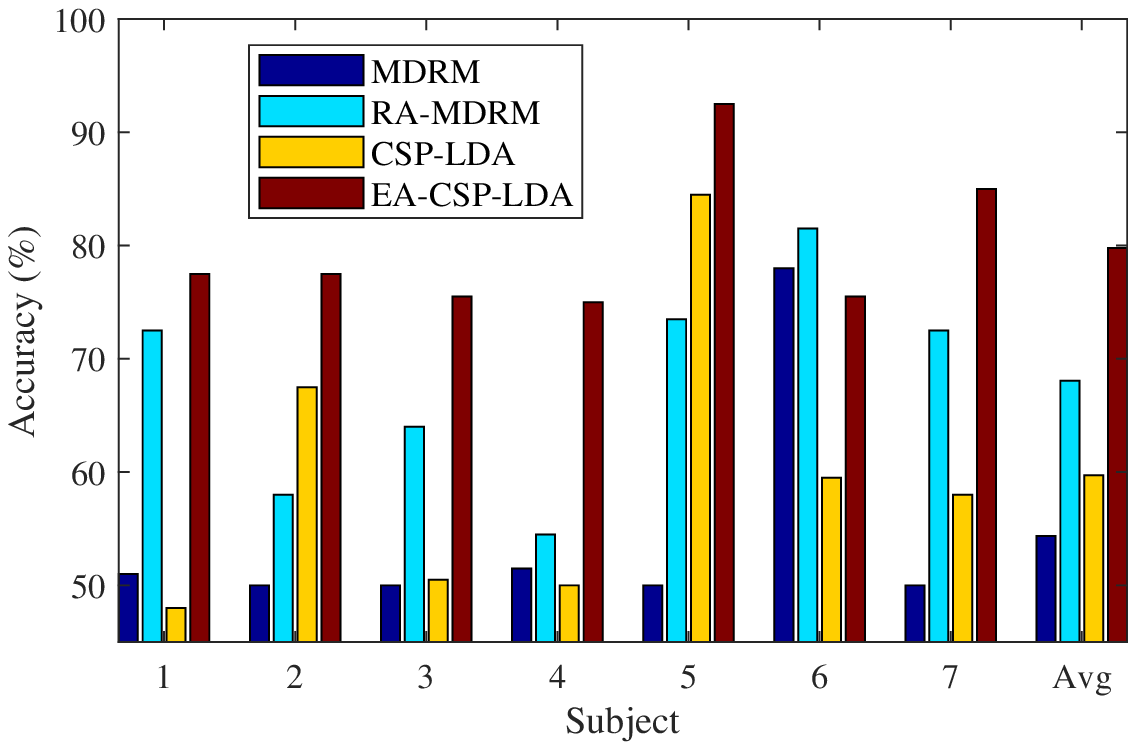}} \label{fig:MI1All}
\subfigure[]{\includegraphics[width=.8\linewidth,clip]{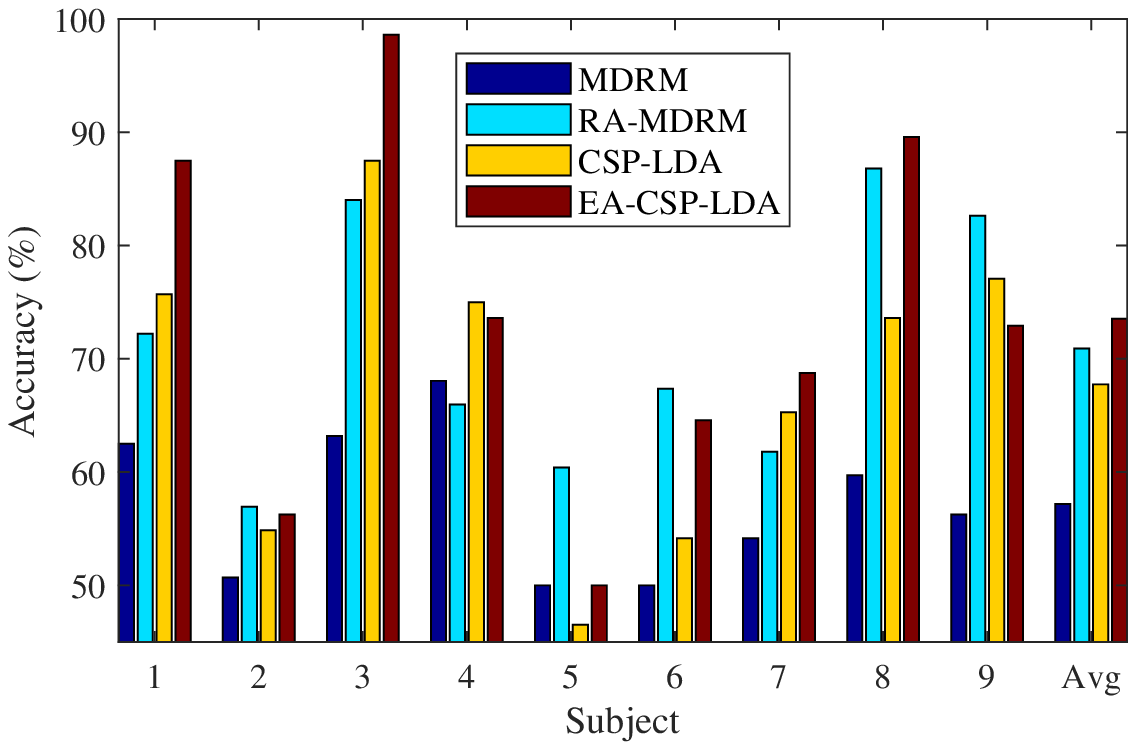}} \label{fig:MI2All}
\caption{Offline unsupervised classification accuracies on the MI datasets: (a) Dataset~1; (b) Dataset~2a. } \label{fig:MI}
\end{figure}

The classification accuracies of the four approaches are presented in Fig.~\ref{fig:MI} and Table~\ref{tab:Acc}, which show that:
\begin{enumerate}
  \item RA-MDRM outperformed MDRM on 15 out of the 16 subjects, suggesting that RA was effective.
  \item EA-CSP-LDA also outperformed CSP-LDA on 14 out of the 16 subjects, suggesting that the proposed EA was also effective.
  \item EA-CSP-LDA outperformed RA-MDRM on 11 out of the 16 subjects, suggesting that the proposed EA, which enables the use of a wide range of Euclidean space signal processing and machine learning approaches, could be more effective than RA.
\end{enumerate}
Finally, it is worth noting that for a small number of subjects (e.g., Subjects 4 and 9 in Dataset 2a), EA actually degraded the classification accuracy. Some possible reasons are explained at the end of the paper, and will be investigated in our future research.

\begin{table}[htpb]\centering \setlength{\tabcolsep}{1mm}
\caption{Offline unsupervised classification accuracies (\%) on the two MI datasets.}\label{tab:Acc}
\begin{tabular}{c|c|cccc} \hline
Dataset&Subject& MDRM & RA-MDRM &CSP-LDA & EA-CSP-LDA  \\ \hline
             &1&51.00&72.50&48.00&\textbf{77.50}\\
             &2&50.00&58.00&67.50&\textbf{77.50}\\
             &3&50.00&64.00&50.50&\textbf{75.50}\\
MI Dataset~1 &4&51.00&54.50&50.00&\textbf{75.00}\\
             &5&50.00&73.50&84.50&\textbf{92.50}\\
             &6&78.00&\textbf{81.50}&59.50&75.50\\
             &7&50.00&72.50&58.00&\textbf{85.00}\\
             &avg&54.36&68.07&59.71&\textbf{79.79}\\\hline
             &1&62.50&72.22&75.69&\textbf{87.50}\\
             &2&50.69&\textbf{56.94}&54.86&56.25\\
             &3&63.19&84.03&87.50&\textbf{98.61}\\
             &4&68.06&65.97&\textbf{75.00}&73.61\\
MI Dataset~2a&5&50.50&\textbf{60.42}&46.53&50.00\\
             &6&50.50&\textbf{67.36}&54.17&64.58\\
             &7&54.17&61.81&65.28&\textbf{68.75}\\
             &8&59.72&86.81&73.61&89.58\\
             &9&56.25&\textbf{82.64}&77.08&72.92\\
             &avg&57.18&70.91&67.75&\textbf{73.53}\\\hline
              \end{tabular}
\end{table}

To determine if the differences between our proposed approach (EA-CSP-LDA) and each other approach was statistically significant, we performed paired-sample $t$-test on the accuracies in Table~\ref{tab:Acc} using MATLAB function \emph{ttest}. The null hypothesis for each pairwise comparison was that the difference between the paired samples has mean zero, and it was rejected if $p \leq \alpha$, where $\alpha = 0.05$ was used. Before performing each $t$-test, we also performed a Lilliefors test \cite{Lilliefors1967} to verify that the null hypothesis that the data come from a normal distribution cannot be rejected.

The paired-sample $t$-test results are shown in Table~\ref{tab:ttestMI}, where the statistically significant ones are marked in bold. EA-CSP-LDA significantly outperformed CSP-LDA on both MI datasets, suggesting that EA was effective. In addition, EA-CSP-LDA significantly outperformed RA-MDRM on Dataset 1, and had comparable performance with it on Dataset 2a, suggesting that EA may be preferred over RA.

\begin{table}[htpb] \centering \setlength{\tabcolsep}{3mm}
\caption{Paired-sample $t$-test results on the test accuracies in Table~\ref{tab:Acc}.}   \label{tab:ttestMI}
\begin{tabular}{c|ccc}   \hline
 \multicolumn{4}{c}{MI Dataset 1}  \\ \cline{1-4}
 & MDRM & RA-MDRM & CSP-LDA \\ \hline
 EA-CSP-LDA & \textbf{0.0030}& \textbf{0.0178}&\textbf{0.0009}\\ \hline \hline
 \multicolumn{4}{c}{MI Dataset 2}  \\ \cline{1-4}
& MDRM & RA-MDRM & CSP-LDA \\ \hline
 EA-CSP-LDA &\textbf{0.0033}& 0.4276 &\textbf{0.0341}\\ \hline
\end{tabular}
\end{table}

It is also interesting to compare the computational cost of different data alignment approaches. The platform was a Dell XPS15 laptop with Intel Core i7-6700HQ CPU@2.60GHz, 16GB memory, and 512 GB SSD, running 64-bit Windows 10 Education and Matlab 2017a. The results are shown in Table~\ref{tab:time}. Our proposed EA-CSP-LDA was 3.6-19.5 times faster than RA-MDRM, and also it had much smaller standard deviation. RA-MDRM ran much slower on Dataset~1 because it had much more channels than Dataset~2a (59 versus 22).

\begin{table}[htpb]\centering  \setlength{\tabcolsep}{3mm}
\caption{The computing time (seconds) of EA-CSP-LDA and RA-MDRM.}\label{tab:time}
\begin{tabular}{c|cc|cc} \hline
& \multicolumn{2}{|c|}{EA-CSP-LDA} & \multicolumn{2}{|c}{RA-MDRM} \\
& Mean & std & Mean & std \\ \hline
MI Dataset~1 &0.3864 &0.0514&7.5326 &0.2200\\ \hline
MI Dataset~2a&0.2405 &0.0322&0.8766 &0.0729\\ \hline
\end{tabular}
\end{table}

In summary, we have demonstrated that our proposed EA is more effective and efficient than RA in offline unsupervised MI classification.

\subsection{Offline Classification Results on the ERP Dataset} \label{sect:OfflineERP}

As RA-MDRM cannot be applied to ERP classification when there are no labeled trials at all from the new subject [RA needs some non-target trials to compute the reference matrix in (\ref{eq:ERPref}), and MDRM needs some target trials to construct $X_i^*$ in (\ref{eq:Xi})], we only validate the effectiveness of EA by comparing it with cases that no data alignment is performed, in leave-one-subject-out cross-validation. All approaches used SVM classifiers, which cannot be associated with RA because RA only outputs covariance matrices.

More specifically, we compared the performances of the following four approaches (all trials were downsampled to 64 Hz):
\begin{enumerate}
\item SVM, which performs principal component analysis (PCA) on the EEG trials to suppress noise and extract features, and then SVM for classification. It does not include any data alignment.
\item EA-SVM, which first performs EA to align the trials from different subjects in the Euclidean space, and then PCA and SVM classification.
\item xDAWN-SVM, which first performs xDAWN \cite{Rivet2009, drwuSF2018} to spatially filter the EEG trials, and then PCA and SVM classification. It does not include any data alignment.
\item EA-xDAWN-SVM, which first performs EA to align the trials from different subjects in the Euclidean space, then xDAWN to spatially filter the EEG trials, and finally PCA and SVM classification.
\end{enumerate}

For all approaches, we first reshaped the 2D features (matrices) of EEG data into 1D vectors, then normalized each dimension to zero mean and unit variance. We then applied PCA to extract 20 features. Because these features had different ranges, we further normalized each feature to interval $[0,1]$. LibSVM \cite{LIBSVM} with a linear kernel was used for classification. We considered the trade-off parameter $C\in\{2^{-3},2^{-2},...,2^5\}$, and used nested 5-fold cross-validation on the training data to identify the optimal $C$. Finally, we used all training data and the optimal $C$ to train a linear SVM classifier, and applied it to the test data.

Because ERPs had significant class imbalance, we used the balanced classification accuracy (BCA) as the performance measure. Let $m^+$ and $m^-$ be the true number of trials from the target and non-target classes, respectively. Let $n^+$ and $n^-$ be the number of trials that are correctly classified by an algorithm as target and non-target, respectively. Then, we first compute
\begin{align}
 a_+=\frac{n^+}{m^+},\quad  a_-=\frac{n^-}{m^-},
\end{align}
where $a_+$ is the classification accuracy on the target class, and $a_-$  on the non-target class. The BCA is then computed as:
\begin{align}
 BCA=\frac{a_++a_-}{2}.
\end{align}

\begin{figure}[htpb]\centering
\includegraphics[width=.8\linewidth,clip]{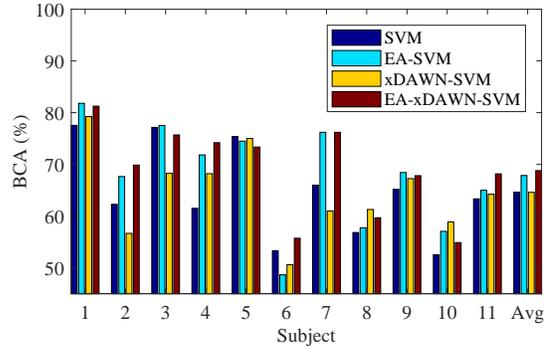}
\caption{BCAs of offline unsupervised classification on the ERP dataset.} \label{fig:RSVPall}
\end{figure}

The BCAs for the four approaches are presented in Fig.~\ref{fig:RSVPall} and Table~\ref{tab:BCA}, which show that:
\begin{enumerate}
\item EA-SVM outperformed SVM on nine out of 11 subjects, suggesting that the proposed EA was generally effective for ERP classification.
\item EA-xDAWN-SVM outperformed xDAWN-SVM on eight out of 11 subjects, suggesting again that the proposed EA was generally effective for ERP classification.
\item On average xDAWN-SVM and SVM achieved similar performances, but EA-xDAWN-SVM slightly outperformed EA-SVM, suggesting that our proposed EA may also help unleash the full potential of xDAWN.
\end{enumerate}

\begin{table}[htpb]\centering \setlength{\tabcolsep}{2mm}
\caption{BCAs (\%) of offline unsupervised classification on the ERP dataset.}\label{tab:BCA}
\begin{tabular}{c|ccccc} \hline
Subject& SVM & EA-SVM &xDAWN-SVM & EA-xDAWN-SVM  \\ \hline
1&77.54&\textbf{81.80}&79.24&81.25\\
2&62.29&67.65&56.67&\textbf{69.86}\\
3&77.16&\textbf{77.52}&68.28&75.70\\
4&61.52&71.83&68.21&\textbf{74.20}\\
5&\textbf{75.38}&74.49&75.00&73.36\\
6&53.27&48.65&50.60&\textbf{55.73}\\
7&65.98&\textbf{76.20}&60.98&\textbf{76.20}\\
8&56.82&57.75&\textbf{61.28}&59.67\\
9&65.18&\textbf{68.43}&67.25&67.82\\
10&52.52&57.06&\textbf{58.87}&54.88\\
11&63.35&64.99&64.23&\textbf{68.17}\\
avg&64.64&67.85&64.60&\textbf{68.80}\\ \hline
\end{tabular}
\end{table}

Paired-sample $t$-tests were also performed for the results in Table~\ref{tab:BCA}. As RA-MDRM could not be applied in this scenario, only two pairs of algorithms were compared, i.e., SVM versus EA-SVM, and xDAWN-SVM versus EA-xDAWN-SVM. The results are shown in Table~\ref{tab:ttestERP}, where the statistically significant ones are marked in bold. EA-SVM significantly outperformed SVM, and EA-xDAWN-SVM significantly outperformed xDAWN-SVM, suggesting that EA was effective on the ERP dataset, too.

\begin{table}[htpb] \centering \setlength{\tabcolsep}{4mm}
\caption{Paired $t$-test results on the test BCAs in Table~\ref{tab:BCA}.}   \label{tab:ttestERP}
\begin{tabular}{c|cc}   \hline
 & SVM & xDAWN-SVM  \\ \hline
 EA-SVM & \textbf{0.0386}& \\ \hline
 EA-xDAWN-SVM && \textbf{0.0449}\\ \hline
\end{tabular}
\end{table}

\subsection{Discussion: Different Choices of the Reference Matrix}\label{sect: reference matrix}

Reference matrix estimation has a direct impact on the performance of the alignment algorithms. RA uses the Riemannian mean of the resting covariance matrices for MI classification, and the Riemannian mean of the non-target covariance matrices for ERP classification [see (\ref{eq:ERPref})]. EA estimates the reference matrix from all trials by (\ref{eq:ref2}), whose procedure is the same for both MI and ERP classification.

In summary, the reference matrix can be estimated from two types of trials for MI classification: 1) the \emph{resting} trials that the subject is not performing any task; and, 2) the \emph{imagery} trials that the subject is performing a motor imagery task. Furthermore, the reference matrix can be computed as the Riemannian mean or the Euclidean mean. So we have four possible combinations: \emph{Riemannian} mean of the \emph{resting} trials (RR), \emph{Euclidean} mean of the \emph{resting} trials (ER), \emph{Riemannian} mean of all \emph{imagery} trials (RI), and \emph{Euclidean} mean of all \emph{imagery} trials (EI).

\begin{figure}[htpb]\centering
\subfigure[]{\label{fig:MI1Ref}\includegraphics[width=\linewidth,clip]{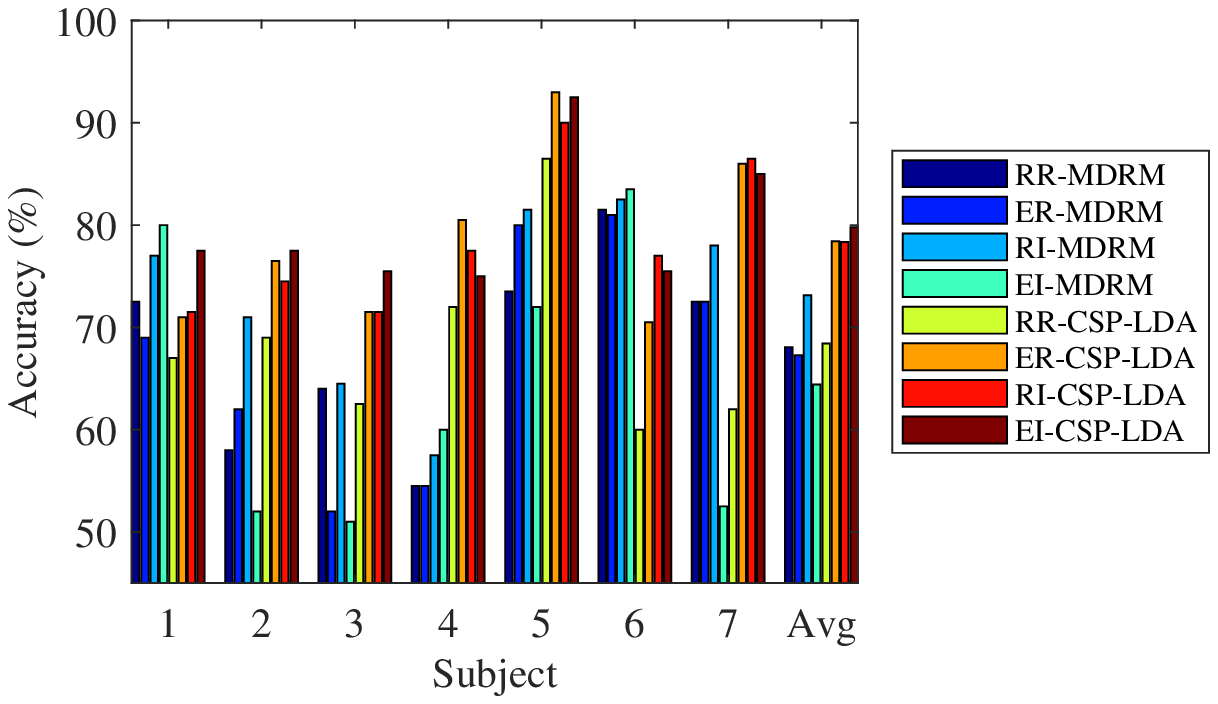}}
\subfigure[]{\label{fig:MI2Ref}\includegraphics[width=\linewidth,clip]{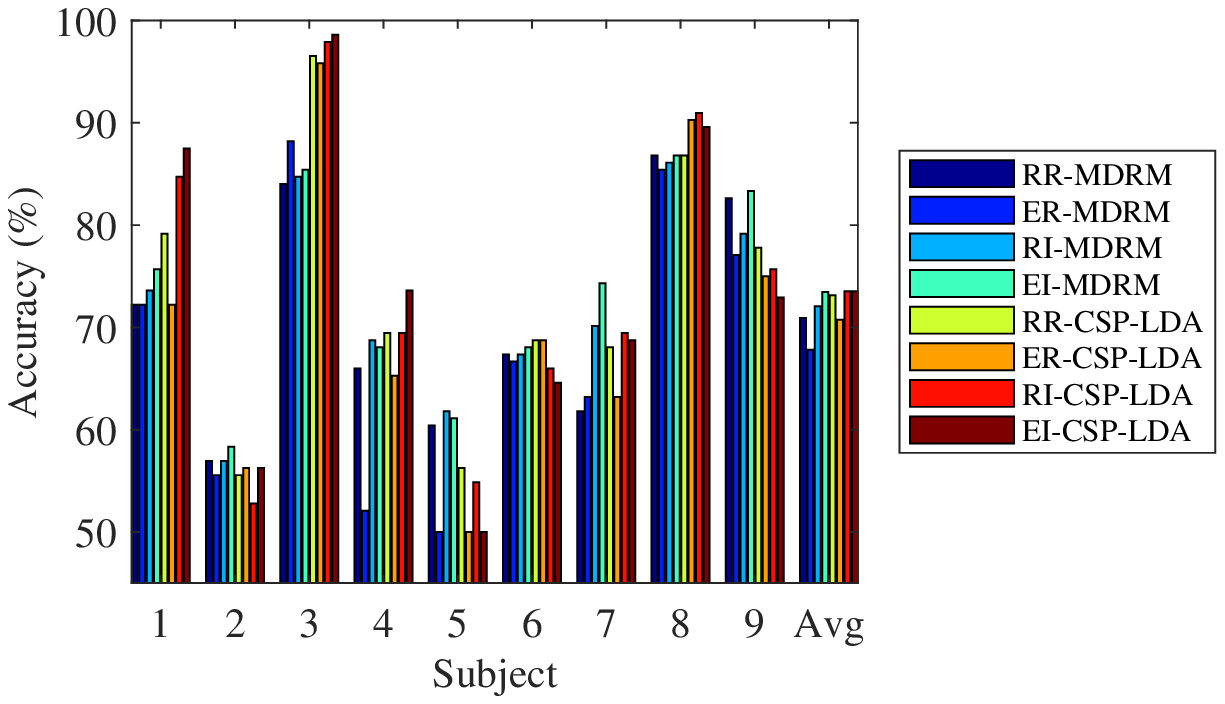}}
\caption{Comparison of different reference matrices on the MI datasets. (a) Dataset~1; (b) Dataset~2a. RR: \emph{Riemannian} mean of the \emph{resting} trials; ER: \emph{Euclidean} mean of the \emph{resting} trials; RI: \emph{Riemannian} mean of all \emph{imagery} trials; EI: \emph{Euclidean} mean of all \emph{imagery} trials.} \label{fig:MIRef}
\end{figure}

This subsection compares the performances of the above four reference matrices. The results are shown in Figs.~\ref{fig:MI1Ref} and \ref{fig:MI2Ref} for MI Datasets~1 and 2a, respectively. They show that:
\begin{enumerate}
\item On average RI-MDRM outperformed RR-MDRM, and EI-CSP-LDA outperformed ER-CSP-LDA, on both datasets, suggesting that estimating the reference matrix from all imagery trials would be better than using all resting trials.

\item On average across all 16 subjects, EI achieved the best performance for CSP-LDA, and RI achieved the best performance for MDRM. This is consistent with our expectation: MDRM operates in the Riemannian space, hence the Riemannian mean might give a more accurate estimation of mean covariance matrices than the Euclidean mean; on the other hand, CSP-LDA operates in the Euclidean space, so the Euclidean mean sounds more reasonable.

\item On average across all 16 subjects, EI-CSP-LDA outperformed RI-MDRM, suggesting that EA was advantageous to RA even when they both used the best reference matrix.
\end{enumerate}

\section{Performance Evaluation: Simulated Online Supervised Classification}\label{sect:Online}

This section evaluates the performance of EA in simulated online supervised classification. The same three datasets were used.

\subsection{Simulated Online Supervised Classification}

In online supervised classification, we have labeled trials from multiple auxiliary subjects, but initially no trials at all from the new subject. We acquire labeled trials from the new subject sequentially on-the-fly, which are then used to train a classifier to label future trials from the new subject, with the help of data from the auxiliary subjects.

We simulated the online supervised classification scenario using the offline datasets presented in Section~\ref{sect:Datasets}. Take MI Dataset~1 as an example. Each time we picked one subject as the new subject, and the remaining six subjects as auxiliary subjects. The new subject had 200 trials. We generated a random integer $n_0 \in [1,200]$, reserved the subsequent $m$ trials $\{n_0+i\}_{i=1}^m$ as the online pool\footnote{When $n_0+i$ was larger than 200, we rewound to the beginning of the trial sequence, i.e., replaced $n_0+i$ by $n_0+i-200$.}, and used the remaining $200-m$ trials as the test data. Starting from an empty training set, we added $r$ trials from the online pool to it each time, built a classifier by combining the training set with the auxiliary data, evaluated its performance on the test data, until all $m$ trials in the online pool were exhausted.

The main difference between offline unsupervised classification and simulated online supervised classification is that the former has a large number of unlabeled trials from the new subject, but none of them have labels, whereas the latter has only a small number of trials from the new subject, all of which are labeled.

\subsection{Simulated Online Classification Results on the MI Datasets}

The four approaches (MDRM, RA-MDRM, CSP-LDA and EA-CSP-LDA) introduced in Section~\ref{sect:OfflineMI} were compared again in simulated online MI classification. In offline unsupervised classification, we had access to all unlabeled EEG trials of the new subject, so its $\bar{R}$ was computed by using all trials for EA, and the resting trials between them for RA. In simulated online supervised classification, we only had access to a small number of labeled trials from the new subject, so its $\bar{R}$ was computed by using these trials for EA, and the resting trials between them for RA (the label information was not needed in either EA or RA; only the EEG trials were used). All labeled trials from auxiliary subjects and the small number of available labeled trials from the new subject were combined to train MDRM, CSP and LDA. We paid special attention to the implementation to make sure it was causal, i.e., we did not make use of EEG and label information that was not supposed to be known at a given time point.

\begin{figure}[htpb]\centering
\subfigure[]{\includegraphics[width=\linewidth,clip]{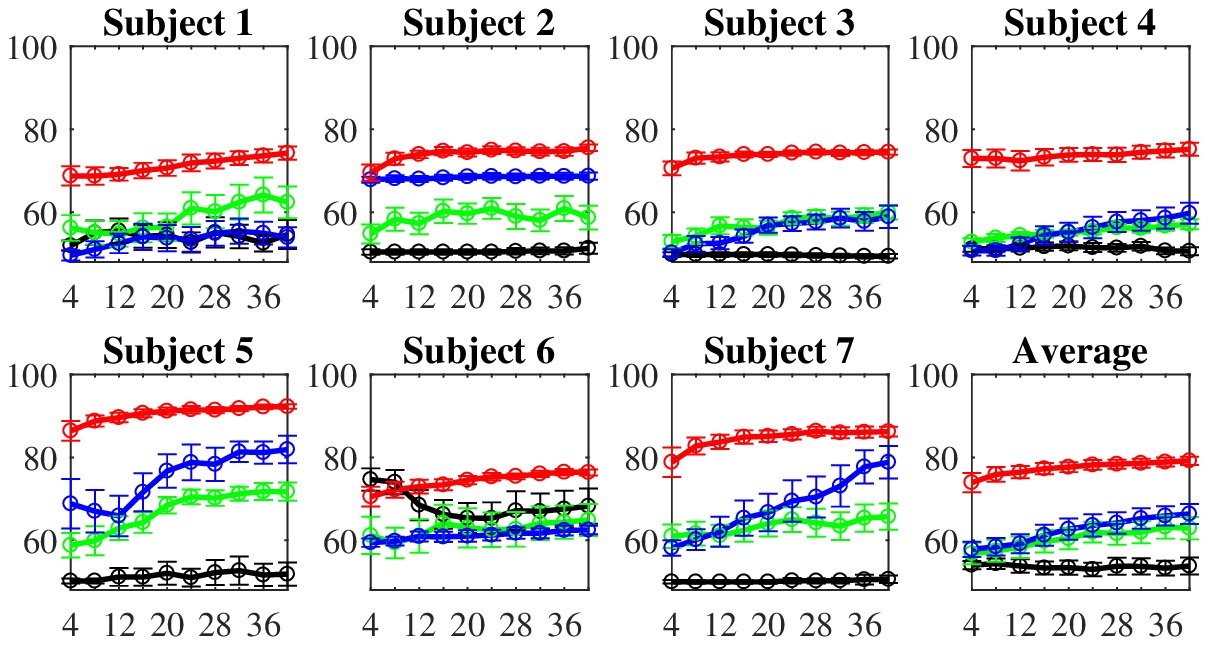}} \label{fig:onlineMI1All}
\subfigure[]{\includegraphics[width=\linewidth,clip]{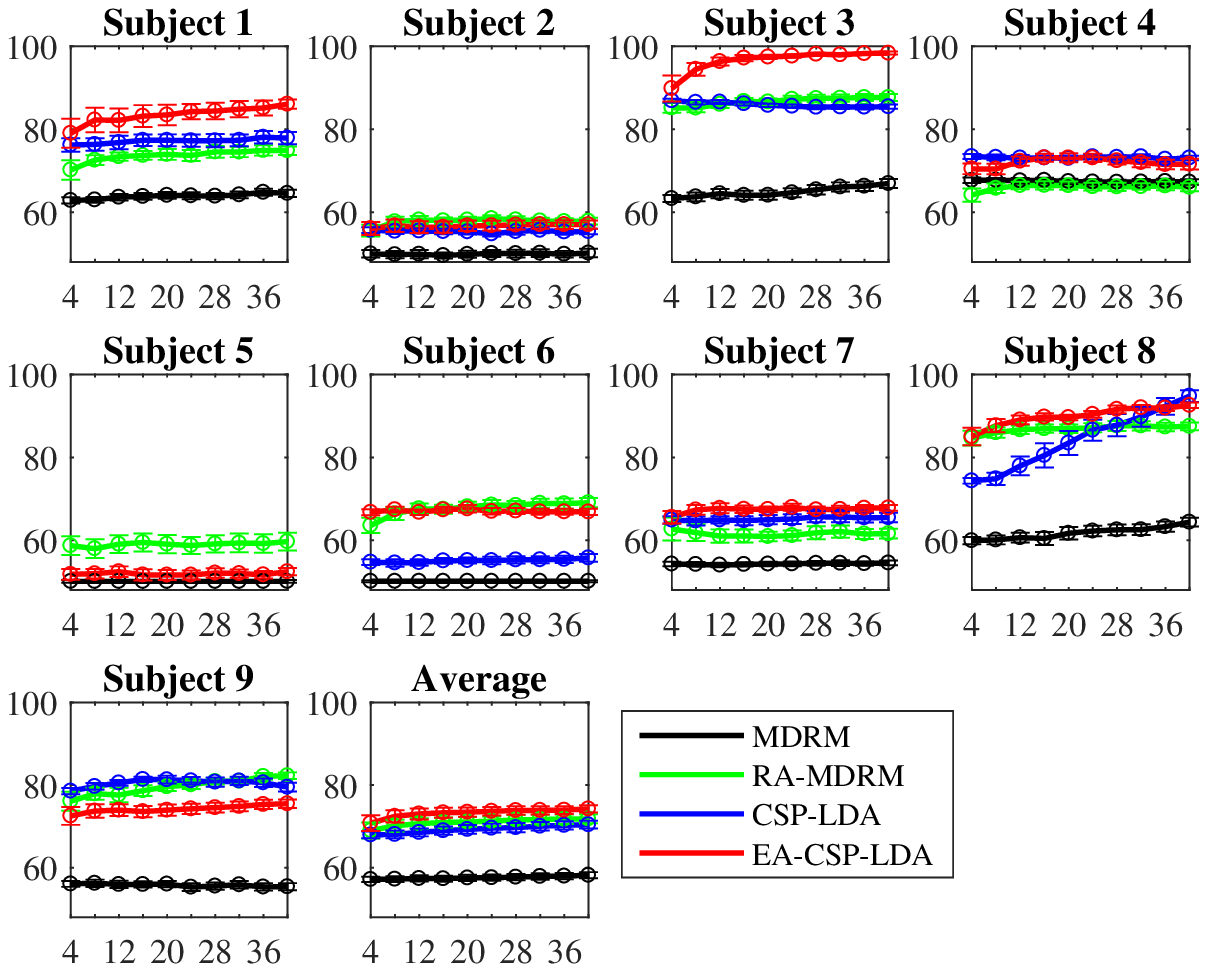}} \label{fig:onlineMI2All}
\caption{Classification accuracies (\%) of simulated online learning on the MI datasets: (a) Dataset~1; (b) Dataset~2a. The horizontal axis shows the number of subject-specific labeled trials from the new subject. The error bars indicate the 95\% confidence intervals. The legends in (a) are the same as those in (b).} \label{fig:onlineMI}
\end{figure}

We used $m=40$ and $r=4$ for both MI datasets. In order to obtain statistically meaningful results, we repeated the experiment 30 times (each time with a random $n_0$) for each new subject. The average classification accuracies of the four approaches are presented in Fig.~\ref{fig:onlineMI}, which shows that:
\begin{enumerate}
  \item RA-MDRM outperformed MDRM on 15 out of the 16 subjects, suggesting that RA was effective in simulated online supervised classification.
  \item EA-CSP-LDA outperformed CSP-LDA on 14 out of the 16 subjects, suggesting that the proposed EA was also effective in simulated online supervised classification.
  \item EA-CSP-LDA outperformed RA-MDRM on 12 out of the 16 subjects, suggesting that EA was generally more effective than RA in simulated online supervised classification.
\end{enumerate}

To determine if the differences between our proposed algorithm and the others were statistically significant in simulated online experiments, we first defined an aggregated performance measure called the area under the curve (AUC). For a particular algorithm on a particular subject, the AUC was the area under its accuracy curve when the number of labeled subject-specific trials increased from 4 to 40. As we repeated the experiments 30 times, we first computed the mean AUC of these 30 repetitions for each subject. Each algorithm had $N$ mean AUCs, where $N$ was the number of subjects. We then compared these mean AUCs using paired-sample $t$-tests. The results are shown in Table~\ref{tab:ttestMIonline}, where the statistically significant ones are marked in bold. EA-CSP-LDA significantly outperformed RA-MDRM on Dataset 1, and had comparable performance with it on Dataset 2a, suggesting that EA may be preferred over RA.

\begin{table}[htpb] \centering \setlength{\tabcolsep}{3mm}
\caption{Paired-sample $t$-test results on the mean AUCs in simulated online MI classification.}   \label{tab:ttestMIonline}
\begin{tabular}{c|ccc}   \hline
 \multicolumn{4}{c}{MI Dataset 1}  \\ \cline{1-4}
 & MDRM & RA-MDRM & CSP-LDA \\ \hline
 EA-CSP-LDA & \textbf{0.0011}& \textbf{0.0001}&\textbf{ 0.0001}\\ \hline \hline
 \multicolumn{4}{c}{MI Dataset 2}  \\ \cline{1-4}
& MDRM & RA-MDRM & CSP-LDA \\ \hline
 EA-CSP-LDA &\textbf{0.0018}& 0.3067 &0.0671\\ \hline
\end{tabular}
\end{table}

\subsection{Simulated Online Classification Results on the ERP Dataset}

Four approaches (MDRM, RA-MDRM, xDAWN-SVM, and EA-xDAWN-SVM) were compared in simulated online supervised classification on the ERP dataset. Note that MDRM and RA-MDRM were not used in offline unsupervised ERP classification because they needed some labeled trials from the new subject to construct the augmented trials, which were not available in offline unsupervised classification. However, they were used in simulated online supervised ERP classification because here labeled trials were available.

\begin{figure}[htpb]\centering
\includegraphics[width=\linewidth,clip]{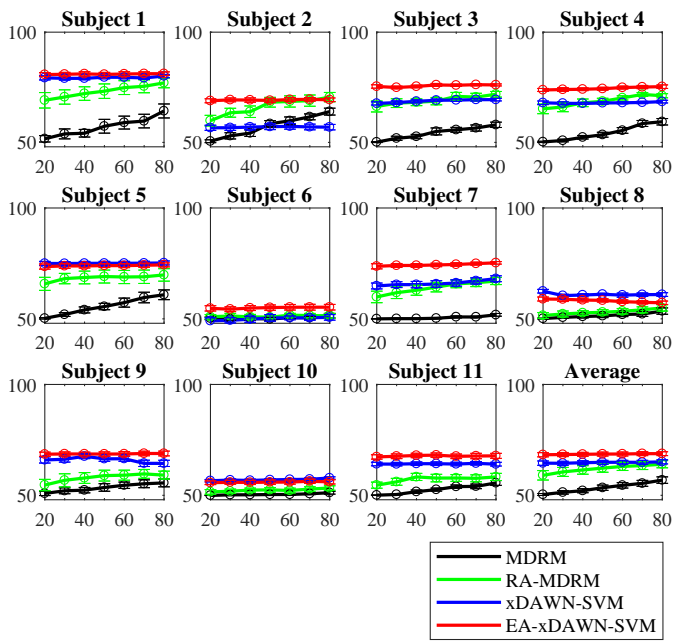}
\caption{BCAs (\%) of simulated online calibration on the ERP dataset. The horizontal axis shows the number of subject-specific labeled trials from the new subject. The error bars indicate the 95\% confidence intervals.} \label{fig:onlineERP}
\end{figure}

We used $m=80$ and $r=10$, and started with 20 trials in the first iteration. In order to obtain statistically meaningful results, we again repeated the experiment 30 times (each time with a random $n_0$) for each new subject. The average BCAs of the four approaches are shown in Fig.~\ref{fig:onlineERP}. Observe that:
\begin{enumerate}
  \item On average RA-MDRM outperformed MDRM, and EA-xDAWN-SVM outperformed xDAWN-SVM, suggesting that both alignment approaches were effective in simulated online supervised classification.
  \item EA-xDAWN-SVM outperformed RA-MDRM on all 11 subjects, suggesting that the proposed EA was more effective than RA in simulated online supervised classification.
\end{enumerate}

Paired-sample $t$-tests were also performed to compare EA-xDAWN-SVM with the other three algorithms. The results are shown in Table~\ref{tab:ttestERPonline}, where the statistically significant ones are marked in bold. EA-xDAWN-SVM significantly outperformed all other approaches, suggesting that the proposed EA was effective and may be preferred over RA.

\begin{table}[htpb] \centering \setlength{\tabcolsep}{3mm}
\caption{Paired-sample $t$-test results on the mean AUCs in simulated online ERP classification.}   \label{tab:ttestERPonline}
\begin{tabular}{c|ccc}   \hline
 & MDRM & RA-MDRM & xDAWN-SVM \\ \hline
 EA-xDAWN-SVM & \textbf{0.0000}& \textbf{0.0000}&\textbf{ 0.0191}\\ \hline
\end{tabular}
\end{table}

\section{Conclusion and Future Research} \label{sect:conclusions}

Transfer learning is a promising approach to improve the EEG classification performance in BCIs, by using labeled data from auxiliary subjects in similar tasks. However, due to individual differences, if the EEG trials from different subjects are not aligned properly, the discrepancies among them may result in negative transfer. A Riemannian space covariance matrix alignment approach (RA) has been proposed to transform the covariance matrices of EEG trials to give them a common reference. However, it has some limitations: 1) it aligns the covariance matrices instead of the EEG trials, so a classifier that operates directly on the covariance matrices must be used to take advantage of the alignment, whereas there are very few such classifiers; 2) its computational cost is high; and, 3) it needs some labeled subject-specific trials from the new subject for ERP-based BCIs.

This paper has proposed a Euclidean space EEG trial alignment approach (EA), which has three desirable properties: 1) it aligns the EEG trials directly in the Euclidean space, and any signal processing, feature extraction and machine learning algorithms can be applied to the aligned trials, so it has much broader applications than the Riemannian space alignment approach; 2) it can be computed several times faster than the Riemannian space alignment approach; and, 3) it does not need any labeled trials from the new subject. Experiments in offline and simulated online classification on two MI datasets and one ERP dataset verified the effectiveness and efficiency of EA.

However, the current EA may still have some limitations. Its goal is to compensate the dataset shift among different subjects, which includes three types of shift:
\begin{enumerate}
\item \emph{Covariate shift} \cite{Shimodaira2000,Sugiyama2008}: the distribution of the inputs (independent variables) changes.
\item \emph{Prior probability shift}: the distribution of the output (target variable) changes.
\item \emph{Concept shift} \cite{Utgoff1986}: the relationship between the inputs and the output changes.
\end{enumerate}
The current EA only considers covariate shift but ignores the other two. So, the per-class input data distributions may still have large discrepancies among different subjects after EA. Moreover, in compensating for the covariate shift, EA may even increase the concept shift, i.e., it is possible that for a specific subject, the two classes become more difficult to distinguish after EA. These could be some of the reasons why EA demonstrated improved performance on most but not all subjects. Another possible reason that EA did not offer advantages on some subjects is that there could be bad trials and/or outliers for these subjects. Including these trials in computing the reference matrix $\bar{R}$ would result in a large error, which further affects the classification accuracy.

Additionally, we acknowledge that the simulated online supervised classification experiments are not identical to real online experiments. Our results would be more convincing if they were obtained from real experiments. Our future research will investigate and accommodate the limitations of EA, and validate the improvements in real-world closed-loop BCI experiments.



\end{document}